\documentclass[runningheads]{llncs}
\usepackage{makecell}
\usepackage{float} 
\usepackage{amsmath}
\usepackage{graphicx}
\usepackage[T1]{fontenc}
\usepackage{graphicx}
\usepackage{array}
\newcolumntype{P}[1]{>{\centering\arraybackslash}p{#1}}

\begin{document}
\title{Heterogeneous graphs model spatial relationship between biological entities for breast cancer diagnosis}
\titlerunning{Heterogeneous graphs model spatial relationship between biological entities}

\author{Akhila Krishna K\inst{1} \and
Ravi Kant Gupta\inst{1} \and
Nikhil Cherian Kurian\inst{1} \and
Pranav Jeevan\inst{1} \and
Amit Sethi\inst{1} }

\authorrunning{A. Krishna et al.}

%
\institute{Indian Institute of Technology Bombay, Mumbai, India \\
\url{https://www.iitb.ac.in/}}

\maketitle              
\begin{abstract}
The heterogeneity of breast cancer presents considerable challenges for its early detection, prognosis, and treatment selection. Convolutional neural networks often neglect the spatial relationships within histopathological images, which can limit their accuracy. Graph neural networks (GNNs) offer a promising solution by coding the spatial relationships within images. Prior studies have investigated the modeling of histopathological images as cell and tissue graphs, but they have not fully tapped into the potential of extracting interrelationships between these biological entities. In this paper, we present a novel approach using a heterogeneous GNN that captures the spatial and hierarchical relations between cell and tissue graphs to enhance the extraction of useful information from histopathological images. We also compare the performance of a cross-attention-based network and a transformer architecture for modeling the intricate relationships within tissue and cell graphs. Our model demonstrates superior efficiency in terms of parameter count and achieves higher accuracy compared to the transformer-based state-of-the-art approach on three publicly available breast cancer datasets -- BRIGHT, BreakHis, and BACH. 
\end{abstract}
\section{Introduction}
Breast cancer is the most common cancer among women globally, and it continues to 
pose significant challenges for early diagnosis, prognosis, and treatment decisions, given its diverse molecular and clinical subtypes~\cite{zielinska2020interaction}. To address these challenges, recent advancements in machine learning techniques have paved the way for improved accuracy and personalized treatment strategies. However, most convolutional neural networks (CNNs) overlook the spatial relationships within histopathological images, treating them as regular grids of pixels~\cite{bai2019nhl,he2016deep,hou2021early,li2019reverse}. To overcome this limitation, graph neural networks (GNNs) have emerged as a promising alternative for classifying breast cancer.

GNNs are designed to handle complex graph structures, making them well-suited for tasks that involve analyzing relationships between entities~\cite{zhou2020graph,wu2020comprehensive,jia2020graphsleepnet}. By representing histopathological images as graphs, with image regions and structures as nodes and their spatial relationships as edges, GNNs can capture the inherent spatial context within the images~\cite{lu2020capturing,raju2020graph,anklin2021learning,chen2020pathomic,zhou2019cgc}. This allows them to extract valuable information and patterns that may be missed by other machine-learning methods.

In this paper, we propose the use of heterogeneous graph convolutions between the cell and the tissue graphs to enhance the extraction of spatial relationships within histology images. This approach allows for the incorporation of diverse, multi-scale, and comprehensive features and spatial relationships by modeling both cell and tissue structures as well as their hierarchical relationships. Specifically, we introduce three features that have not been previously used in GNNs for histopathology images: (1) heterogeneous graph convolutions along with a transformer which outperforms~\cite{hou2022spatial} on BRIGHT~\cite{brancati2022bracs}, (2) an adaptive weighted aggregation technique with heterogeneous convolutions that outperforms~\cite{hou2022spatial} and is more efficient in terms of the number of parameters, and (3) the cross-attention modules of CrossVit~~\cite{chen2021crossvit} along with heterogeneous graph convolutions to extract the spatial relationship between cell and tissue graphs. We also analyzed different methods of k-nearest neighbor (kNN) graph building for cell and tissue graphs and found that edges based on node feature similarities performed better than other methods such as spatial closeness~~\cite{pati2020hact,pati2022hierarchical} or dynamically learnable layers for edge creation~\cite{hou2022spatial}. Extensive experiments conducted on three publicly available breast cancer histology datasets -- BRIGHT~~\cite{brancati2022bracs}, BACH~~\cite{aresta2019bach}, and BreakHis~~\cite{benhammou2020breakhis} -- demonstrate the gains of our method.



\section{Related Work}
The first work on entity-level analysis for histology was~\cite{zhou2019cgc} in which a cell graph was created and a graph convolutional network was used for processing the graph.  Other works, such as~\cite{pati2020hact,pati2022hierarchical,hou2022spatial}, have also focused on the entity-level analysis of histology images by constructing cell graphs and tissue graphs and by constructing more than two levels of subgraphs to capture the spatial relationship in the image. In~\cite{pati2020hact}, the authors introduced an LSTM-based feature aggregation technique to capture long-range dependencies within the graphs. By utilizing the LSTM, they aimed to capture the temporal dependencies between cells and tissues, enhancing the understanding of their relationships.
In contrast,~\cite{hou2022spatial} took a different approach by constructing more than two levels of subgraphs and using a spatial hierarchical graph neural network. This network aimed to capture both long-range dependencies and the relationships between the cell and tissue graphs. To achieve this, the authors employed a transformer-based feature aggregation technique, leveraging the transformer's ability to capture complex patterns and dependencies in the data. However, it was observed that both of these approaches fell short of fully extracting the intricate relationship between tissue and cell graphs. Hence, there is a scope to explore alternative methods or combinations of techniques to better understand and leverage the spatial relationships within histology images for improved analysis and classification of breast cancer and other diseases. 
\section{Methodology}
\begin{figure}
\includegraphics[scale=0.17]{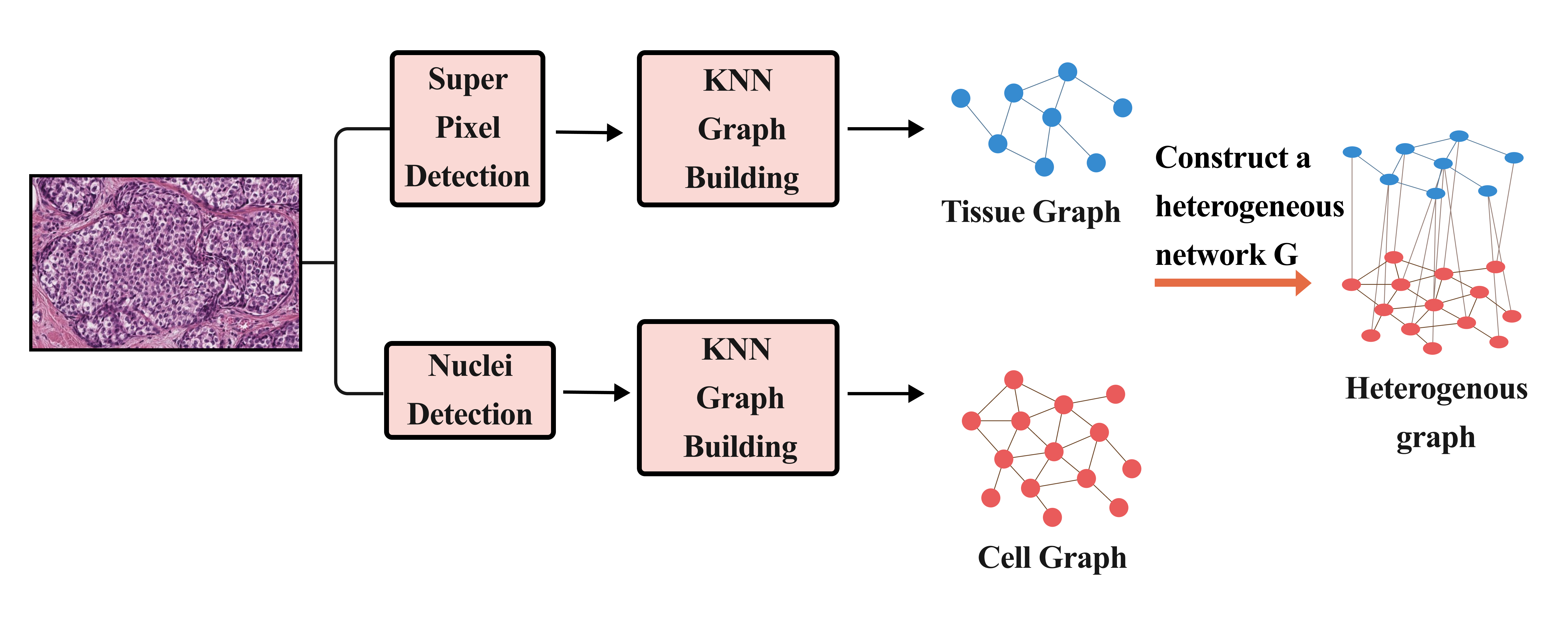}
\caption{Cell and tissue graph formation } \label{fig1}
\end{figure}
\subsection{Tissue and cell graph extraction from histology images}
The interrelationship between cells and tissue is investigated through the construction of cell and tissue graphs. The goal is to perform multi-level feature analysis by extracting relevant features from these entities. To construct the cell graph, a pre-trained model called HoverNet~\cite{graham2019hover} is used for nuclei detection. The feature representation for each nucleus is extracted by processing patches around the nuclei with a ResNet34 encoder~\cite{he2016deep,hou2022spatial}. Similarly, the tissue entities are identified using segmentation masks, which are obtained by applying the Simple Linear Iterative Clustering (SLIC) algorithm for superpixel segmentation~\cite{achanta2012slic,hou2022spatial}. Once the tissue entities are identified, their feature representations are extracted in the same manner as the cell entities. We then utilize kNN to get the $k$ most similar nodes of all nodes based on the distance of node feature representations for the formation of edges between cell entities and tissue entities. In our experiment, we used $k=5$ for all models and datasets. The edges between cells and tissues are formed by using the spatial position of cells and tissues. We treat a cell node $C_i$ and a tissue node $T_i$ as connected if $C_i\in T_i$.

\subsection{Heterogeneous Graph Convolution}
The interaction between cell and tissue graphs has to be captured effectively for better analysis which is done using heterogeneous graph convolutions. We define a heterogenous graph($H$) as a union of cell-to-cell, tissue-to-tissue, and cell-to-tissue relations (which are defined by edges) along with their node features.
\begin{equation}
    H = {\{C,T, E_{cell->cell}, E_{tissue->tissue}, E_{cell->tissue}\}}
\end{equation}
where $C$ and $T$ are the features of the node in the cell graph and the tissue graph and $E_{A->B}$ is the list of edges between the elements in sets A and B. 

We utilize Graph Sage convolutions\cite{hamilton2017inductive} to transmit messages individually from the source node to the target node for each relation. When multiple relationships direct to the same target node, the outcomes are combined using a specified aggregation method. This is how heterogeneous graph convolution is implemented. We can formulate it as follows: let $z_{jR_i}$ be the output vector representations on node $j$ due to Graph Sage convolution on the nodes defined by relation $R_i$, where $R_i\in \{ cell->cell, cell->tissue, tissue->tissue \}$, then the final vector representation on node $j$, $z_j$ is 
\begin{equation}
    z_j = Aggregator(\{z_{jR_u}, \forall u \in U\} )
\end{equation}
where $U$ is the set of relations directing to the node $j$. This is implemented using PyTorch geometric library.

\begin{figure}
\centering
\includegraphics[scale = 0.05]{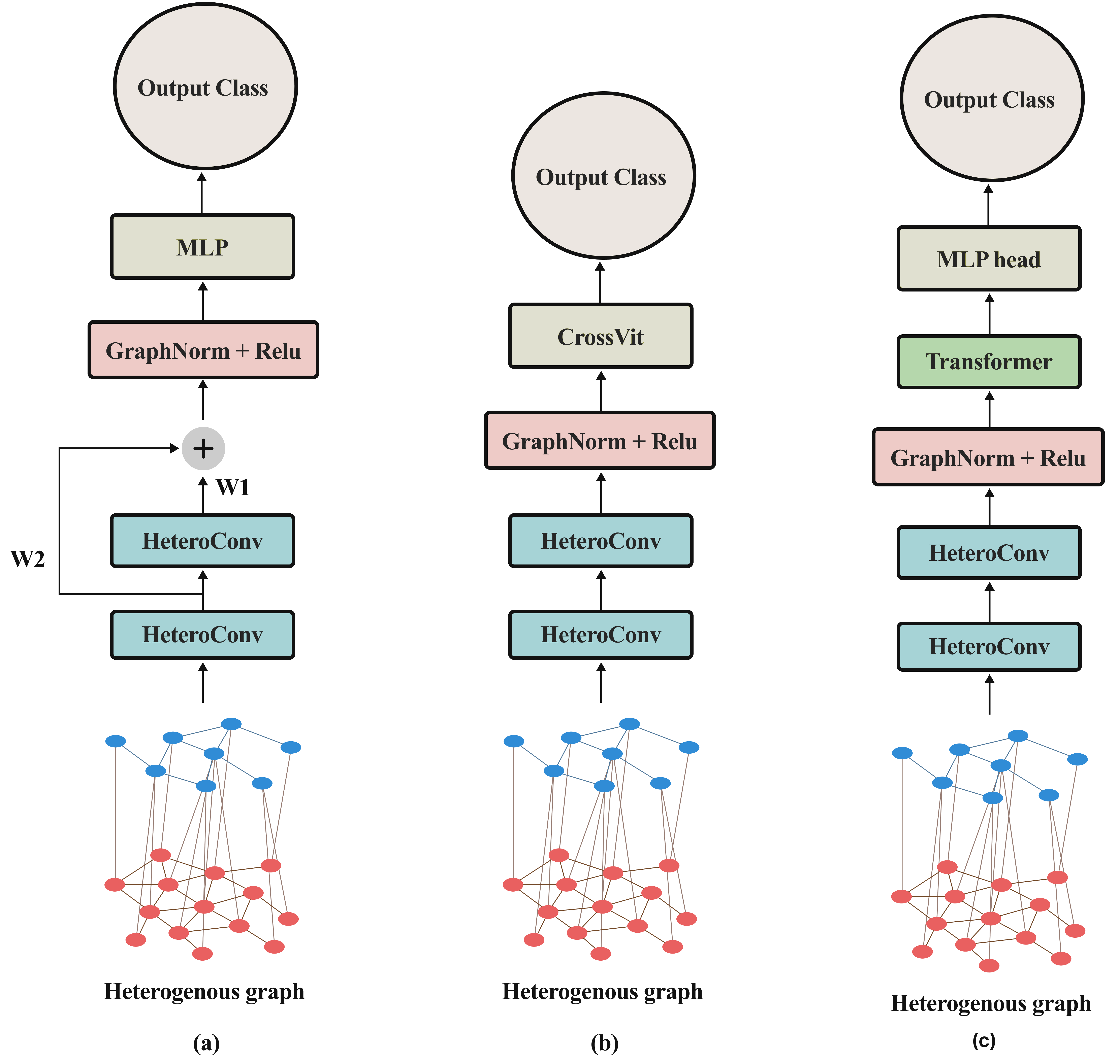}
\caption{Proposed architectural variants of Heterogeneous Graph Neural Network (HG): (a) HG with adaptive weighted aggregation of multi-level features, (b) HG with CrossVit for hierarchical feature fusion, and (c) HG with transformer for attentive interaction.} \label{fig1}
\end{figure}
\subsection{Adaptive weighted aggregation for multi-level feature fusion}
Low-level and high-level features are important in classification tasks. So we use skip connections for its extraction and adaptive weighted aggregation layer for feature fusion. Adaptive weighted aggregation layer $f$ is represented as follows : 
\begin{equation}
f(w_1,w_2,...w_n,F_1,F_2....F_n) = w_1F_1+w_2F_2+w_3F_3+.......w_nF_n,
\end{equation}
where $w_i$ is trainable weights of dimension $1\times1$ and $F_i$ is feature vector of dimension $256\times1$  for $i\in {1,2,3,...n}$.

\subsection{Cross-attention feature fusion and attentive interaction using transformers}
The transformer encoder of~\cite{vaswani2017attention} which uses self-attention is used for extracting the long-range dependencies between the nodes in the graph. Formally, let us say the transformer encoder is $T$ and the input to $T$ is $Concat(N_c, N_t)$ where $N_c$ and $N_t$ are node features of cell and tissue graph respectively. The output from the transformer encoder goes to MLP layers to get the final output. $N_c$ and $N_t$ are of dimension $C\times1\times256$ and $C\times1\times256$ respectively where $C$ is the number of nodes in the cell graph (we extrapolate the tissue graph with zeros for expanding the dimension of it to the same dimension as that of cell graph).

Cross-attention modules of~\cite{chen2021crossvit} are used for extracting the interaction between cell graph and tissue graph which then is then fed into multi-layer perceptron layers to get the required output.

\section{Experiments and Results}

\subsection{Clinical Datasets and evaluation methods}

For assessing the utility of graph based deep learning methods, breast cancer histology datasets form an ideal test bed due to the diversity of its subtypes. We used three datasets in this work. The BRIGHT dataset is a breast cancer subtyping dataset released as a part of the Breast tumor Image classification on Gigapixel Histopathological images challenge~\cite{brancati2022bracs}. The breast cancer histopathology image datasets (BACH)~\cite{aresta2019bach} has four disease states -- normal, benign, ductal carcinoma in-situ (DCIS), and invasive ductal carcinoma (IDC). The BreakHis~\cite{benhammou2020breakhis} has malignant and non-malignant classes.

\subsubsection{BRIGHT Dataset}

The BRIGHT dataset contains 4025 hematoxylin \& Eosin (H\&E) stained breast cancer histology images scanned at 0.25mn/pixel resolution. It is classified into 6 classes: Pathological Benign (PB), Usual Ductal Hyperplasia (UDH), Flat Epithelial Atypia (FEA), Atypical Ductal Hyperplasia (ADH), Ductal Carcinoma In Situ (DCIS) and Invasive Carcinoma (IC). These are further grouped into 3: cancerous tumors (DCIS and IC), non-cancerous tumors (PB and UDH) and pre-cancerous tumors (FEA and ADH)~\cite{brancati2022bracs}.

\subsubsection{BACH Dataset}
The BACH dataset contains 400  hematoxylin and eosin (H\&E) stained breast cancer histology images with pixel scale $0.42\mu m \times 0.42\mu m $. It is classified into 4 classes: normal, benign, in situ carcinoma, and invasive carcinoma~\cite{aresta2019bach}.

\subsubsection{BreakHis dataset}
The BreakHis dataset contains 9109 microscopic images of breast cancer at 400X magnifying factor. It contains 2480 benign and 5429 malignant tissue images~\cite{benhammou2020breakhis}.

\subsection{Experimental Setup} 

All the experiments are implemented in Pytorch and using Pytorch-geometric and histocartography library~\cite{jaume2021histocartography}. The F-score was used as the evaluation metric. We proposed three models and compared each other and also compared with the state-of-the-art model~\cite{hou2022spatial}. 
Hovernet was used for nuclei detection in all models for all datasets. The feature vector of length 512 for each nucleus was extracted from a patch size of 72, 72, and 48  around each nuclei using Resnet 34 model~\cite{he2016deep} for the BRIGHT, BACH, and BreakHis respectively. We also compared 3 methods of edge formation of graphs: dynamic structural learning~\cite{hou2022spatial}, kNN algorithm on the distance between cell entities and the tissue entities, and kNN algorithm on node feature representation of cell entities and tissue entities. The batch size was set to 32. A learning rate of $1e^{-4}$ and Adam optimizer with a weight decay of $5e^{-4}$ were used while training the models. 

\subsection{Comparison with the state-of-the-art methods}
The results for six-class classification and three-class classification on the BRIGHT dataset are listed in Table \ref{tab1} and \ref{tab2} respectively. The four-class classification on the BACH dataset is shown in Table \ref{tab3} and the 2-class classification on the BreakHis dataset is shown in Table \ref{tab4}.



\begin{table}
\centering
\renewcommand{\arraystretch}{1.2}
\setlength{\tabcolsep}{1.6pt}
\caption{Weighted F-score (\%) of HG compared with other methods on BRIGHT dataset for six-class classification.}\label{tab1}
\begin{tabular}{|c|P{0.9cm}|P{0.8cm}|P{0.8cm}|P{0.9cm}|P{0.9cm}|P{0.9cm}|c|c|}
\hline
Model &  DCIS & IC & PB & UDH & ADH & FEA & \makecell{Weighted\\ F-score} & \makecell{Number of parameters \\ (in million)}\\
\hline
SHNN & 72.9 & 87.3 & 74.7 & 51.8 & 45.0 & 73.9 & 69.6 & 2.2 \\
\hline
HG + AWA & 72.8 & 88.0 & 72.7 & \bf{54.1} & 48.9 & 74.2 & 70.1 & \bf{1.5} \\
\hline
HG + CrossVit & 77.9 & 88.0 & 71.1 & 53 & \bf{54} & 73 & 71.3 & 2.5\\
\hline
HG+transformer & \bf{84} & \bf{92} & \bf{79} & 54 & 49 & \bf{78.5} & \bf{75.3} & 2.0\\

\hline
\end{tabular}
\end{table}

\begin{table}
\centering

\renewcommand{\arraystretch}{1.2}
\setlength{\tabcolsep}{1pt}

\caption{Weighted F-score (\%) of HG compared with other methods  on BRIGHT dataset across four test folds for three-class classification.}\label{tab2}
\begin{tabular}{|c|c|c|c|c|c|c|c|c|}
\hline
Model & 1 & 2 & 3 & 4 & $\mu\pm\sigma$ & Cancerous & Non-cancerous & Pre-cancerous\\
\hline
SHNN & 78 & 78.11 & 78.93 & 79.21 & 78.56$\pm$0.56 & 85.21$\pm$2.07 & 78.08$\pm$0.81 & 72.11$\pm$0.82 \\
\hline
HG + CrossVit & 78.75 & 78.28 & 77.2 & 74.29 & 77.13$\pm$2.00 & 83.93$\pm$1.12 & 76.29$\pm$3.08 & 70.38$\pm$2.35 \\
\hline
HG + AWA & 80.24 & 77.98 & 80.82 & \bf{80.55} & 79.89$\pm$1.30 & 87.08$\pm$1.49 & 78.09$\pm$2.64 & 73.75$\pm$1.36\\
\hline
HG+transformer & \bf{83.45} & \bf{82.59} & \bf{81.86} &  80.36 & \bf{82.06$\pm$1.31} & \bf{88.78$\pm$1.96} & \bf{80.61$\pm$1.33} & \bf{76.08 $\pm$1.36} \\

\hline
\end{tabular}
\end{table}

\begin{table}

\centering
\renewcommand{\arraystretch}{1.4}
\setlength{\tabcolsep}{2.1pt}
\caption{Weighted F-score (\%) of HG compared with other methods  on BACH dataset.}\label{tab3}
\begin{tabular}{|c|c|c|c|c|c|}

\hline
Model &  Normal & Benign & InSitu & Invasive & \makecell{Weighted F-score}\\
\hline
SHNN & \bf{90} & 80 & 85.10 & 84.21 & 84.83\\
\hline
\makecell{HG+transformer} & 88.88 & \bf{86.48} & \bf{85.20} & \bf{90.10} & \bf{86.62} \\

\hline
\end{tabular}
\end{table}

\begin{table}[H]

\centering
\renewcommand{\arraystretch}{1.4}
\setlength{\tabcolsep}{2.1pt}
\caption{Weighted F-score (\%) of HG compared with other methods  on BreakHis dataset.}\label{tab4}
\begin{tabular}{|c|c|c|c|c|c|}
\hline
Model &  Malignant & Benign  & \makecell{Weighted F-score}\\
\hline
SHNN & 89.16 & 78.17 & 85.46\\
\hline
\makecell{HG+transformer} & \bf{95.62} & \bf{91.70} & \bf{94.30} \\

\hline
\end{tabular}

\end{table}


\subsection{Ablation Studies}
We performed studies on different methods of graph formation and removing and adding different layers of the model. The results are listed in Table~\ref{tab5}. 
The spatial hierarchical neural network (SHNN) of~\cite{hou2022spatial} is Graph Sage convolutions in homogeneous graphs with a transformer. It can be observed that the addition of a transformer to Graph Sage convolutions in the SHNN led to an increase in F-score by 4\% whereas in the case of a heterogeneous convolutional network the addition of a transformer to heterogeneous convolutions (HG) gave an increase in the F-score of nearly 7\%. It can be concluded that heterogeneous convolutions led to an increase in interactions between cell and tissue features such that the output of these convolutions has better features for the transformer to work on than that of normal sage convolutions. It is also to be noted that in our model we used just two heterogeneous convolutional layers as opposed to 6 graph sage convolutional layers in~\cite{hou2022spatial}.

We tried three different graph formation methods for cell and tissue graph formation. The results are listed in Table~\ref{tab6}. Our method of graph formation using kNN algorithm on node features was seen to perform better than all of the other three methods.

\begin{table}
\centering
\renewcommand{\arraystretch}{1.2}
\setlength{\tabcolsep}{1.2pt}
\caption{Weighted F-score (\%) of different graph-based methods compared on BRIGHT dataset.}\label{tab5}
\begin{tabular}
{|c|P{0.9cm}|P{0.8cm}|P{0.8cm}|P{0.9cm}|P{0.9cm}|P{0.9cm}|c|}
\hline
Model &  DCIS & IC & PB & UDH & ADH & FEA & Weighted F-score\\
\hline
Graph Sage Conv  & 72.1 & 84.8 & 70 & 39 & 44.2 & 68.5 & 65.3\\
\hline
SHNN  & 72.9 & 87.3 & 74.7 & 51.8 & 45 & 73.9 & 69.6\\
\hline
HG  & 76.5 & 85.6 & 72.5 & 49.1 & 40.25 & 72 & 68.2\\
\hline
HG+transformer & 84 & 92 & 79 & 54 & 49 & 78.5 & 75.3\\

\hline
\end{tabular}
\end{table}

\begin{table}
\centering
\renewcommand{\arraystretch}{1.2}
\setlength{\tabcolsep}{1.2pt}
\caption{Weighted F-score (\%) of different edge-formation methods compared on BRIGHT dataset.}\label{tab6}
\begin{tabular}{|c|P{0.9cm}|P{0.8cm}|P{0.8cm}|P{0.9cm}|P{0.9cm}|P{0.9cm}|c|}
\hline
Model &  DCIS & IC & PB & UDH & ADH & FEA & Weighted F-score\\
\hline
DSL  & 81.7 & 90.9 & 78.0 & 55.0 & 48.3 & 75.2 & 73.6\\
\hline
KNN on distance & 81.1 & 90.4 & 77.8 & 56.8 & 57.7 & 75.9 & 74.9\\
\hline
KNN on node features& 84.0 & 92 & 79 & 54 & 49 & 78.5 & 75.3\\

\hline
\end{tabular}
\end{table}

\section{Conclusion}

We introduced three novel architectures for histopathological image classification based on heterogeneous graph neural networks. Our research highlights the capability of heterogeneous graph convolutions in capturing the spatial as well as hierarchical relationship within the images, which allows it to surpass the performance of existing methods for histopathology image analysis. This emphasizes the significance of considering the relationship between cells and the surrounding tissue area for accurate cancer classification. Additionally, we established that the self-attention-based model outperforms the cross-attention-based model. We attribute this observation to the ability of self-attention to extract long-range dependencies within the graphs. Furthermore, we demonstrated the importance of analyzing the relationship between similar parts of the histopathology image, showcasing that constructing the graphs based on the similarity between node features yields superior results compared to other approaches, such as those based on spatial distance. However, we think that in the future both similarity and spatial distance should be combined for graph edge formation. It would also be good to explore novel techniques for enhancing graph convolutions in order to extract long-range dependencies within the graph more effectively. Additionally, there is a potential to develop innovative methods for graph pooling that minimize information loss. These directions of research would contribute to further advancements in histopathological image classification using heterogeneous graph neural networks.

\newpage
%
%
\nocite{*}
\bibliographystyle{splncs04}
\bibliography{references}





\end{document}